\documentstyle[colap]{article}

\newcommand{\avmplus}[1]{{\setlength{\arraycolsep}{0.4mm}	
                       \renewcommand{\arraystretch}{0.7}
                       \left[ 			
                       \begin{array}{l}
                       \\[-2mm] #1 \\[-2mm] \\
                       \end{array} 		
                       \right]
                    }}

\newcommand{\attval}[2]{{\mbox{\scriptsize {\sc #1}}\! =\!\! {{#2}}}}
\newcommand{\attvaltyp}[2]{{\mbox{\scriptsize{\sc #1}}\! =\! {\myvaluebold{#2}}}}

\newcommand{\myvaluebold}[1]{{\mbox{\scriptsize {\bf #1}}}}

\newcommand{\ind}[1]{{\setlength{\fboxsep}{0.5mm} \fbox{{\scriptsize #1}} \!}}

\newtheorem{exampl}{Ex.}
\newenvironment{example}{\begin{quote}\footnotesize \begin{exampl}%
\begin{minipage}[t]{2in}}
{\end{minipage}\end{exampl}\end{quote}}

\newtheorem{assume}{Assumption}
\newtheorem{definition}{Definition}
\newtheorem{condition}{Condition}

\newtheorem{proof}{Proof}

\newenvironment{exquote}{\begin{quote}\footnotesize}
{\end{quote}}

\newcommand{\SetFigFont}[3]{\footnotesize \csname#3\endcsname}

\newcommand{\tab}{\hspace*{3em}}

\newcommand{\smallnl}{\\[0.5em]}

\newenvironment{sdescription}%
{\begin{list}{}{\setlength{\itemsep}{-0.5em}
\setlength{\topsep}{0em}}}
{\end{list}}

\def\pterror#1{\errmessage{Parsetree ERROR: #1}}

\newdimen\pthgap\def\pthorgap#1{\pthgap=#1}
\newdimen\ptvgap\def\ptvergap#1{\ptvgap=#1}

\newbox\ptnodestrutbox\def\ptnodestrut{\unhcopy\ptnodestrutbox}
\newbox\ptleafstrutbox\def\ptleafstrut{\unhcopy\ptleafstrutbox}

\def\ptnodefont#1#2#3{\def\ptnodefn{#1}
  \setbox\ptnodestrutbox=\hbox{\vrule height#2 width0pt depth#3}}
\def\ptleaffont#1#2#3{\def\ptleaffn{#1}
  \setbox\ptleafstrutbox=\hbox{\vrule height#2 width0pt depth#3}}

\pthorgap{12pt}                         
\ptvergap{12pt}                         
\ptnodefont{\normalsize\rm}{11pt}{3pt}  
\ptleaffont{\normalsize\it}{11pt}{3pt}  

\newcount\ptdepth           
\newcount\ptn               
\newbox\ptm \newdimen\ptmx  
\newbox\pta \newdimen\ptax  
\newbox\ptb \newdimen\ptbx  
\newbox\ptc \newdimen\ptcx  
\newbox\ptx \newdimen\ptxx  
\newif\ifpttri              

\def\ptnext{\advance\ptn by 1 \ifcase\ptn
  \or \setbox\ptm=\box\ptx \ptmx=\ptxx \or \setbox\pta=\box\ptx \ptax=\ptxx
  \or \setbox\ptb=\box\ptx \ptbx=\ptxx \or \setbox\ptc=\box\ptx \ptcx=\ptxx
  \else \pterror{More than 3 daughters in (sub)tree}\fi}

\def\ptbegtree{\ptdepth=0}
\def\ptendtree
  {\ifnum\ptdepth>0 \pterror{Mismatched bracketing: too few ')'s!}\fi}

\def\ptbeg{\ifnum\ptdepth=0 \leavevmode\fi\begingroup
  \advance\ptdepth1 \ptn=0\pttrifalse}
\def\ptend{\ifnum\ptdepth=0 \pterror{Mismatched bracketing: too many ')'s!}
  \else\ptcons\endgroup\ifnum\ptdepth=0 \box\ptx\else\ptnext\fi\fi}

\def\ptnodeaux#1{\setbox\ptx=\hbox{#1}\ptxx=0.5\wd\ptx\ptnext}
\def\ptnode#1{\ptnodeaux{\ptnodefn\ptnodestrut #1}}
\def\ptleaf#1{\ptnodeaux{\ptleaffn\ptleafstrut #1}}

\def\ptcons
 {\ifnum\ptn=0 \ptconsz 
  \else
    \ifpttri\ptconstri\else
      \ifcase\ptn \or\ptconsm\or\ptconsma\or\ptconsmab\or\ptconsmabc \fi \fi
    \ptax=\ptxx \advance\ptax-\ptmx \ptbx=0pt
    \ifdim\ptax<0pt \ptbx=-\ptax\ptax=0pt\ptxx=\ptmx \fi
    \setbox\ptx=\vtop{\hbox{\kern\ptax\box\ptm}\nointerlineskip
                      \hbox{\kern\ptbx\box\ptx}}\fi
  \global\ptxx=\ptxx\global\setbox\ptx=\box\ptx}

\def\ptavg#1#2#3{#1=#2\advance#1#3#1=0.5#1}     
\def\ptadv#1#2{\advance#1#2\advance#1\pthgap}   

\def\ptconsz{\ptxx=0pt \setbox\ptx=\vtop{}}     

\def\ptconsm{\ptxx=0pt 
  \setbox\ptx=\hbox{\ptedge{1}{0}{}{}}}         

\def\ptconsma                                   
 {\ptxx=\ptax\setbox\ptx=\vtop{
    \hbox{\ptedge{1}\ptax{}{}}\nointerlineskip
    \hbox{\box\pta}}}

\def\ptconsmab                                  
 {\ptadv\ptbx{\wd\pta}\ptavg\ptxx\ptax\ptbx
  \setbox\ptx=\vtop{
    \hbox{\ptedge{2}\ptax\ptbx{}}\nointerlineskip
    \hbox{\box\pta\kern\pthgap\box\ptb}}}

\def\ptconsmabc                                 
 {\ptadv\ptbx{\wd\pta}\ptadv\ptcx{\wd\pta}%
  \ptadv\ptcx{\wd\ptb}\ptavg\ptxx\ptax\ptcx
  \setbox\ptx=\vtop{
    \hbox{\ptedge{3}\ptax\ptbx\ptcx}\nointerlineskip
    \hbox{\box\pta\kern\pthgap\box\ptb\kern\pthgap\box\ptc}}}

\def\ptconstri                                  
 {\ifcase\ptn\or\setbox\pta\hbox{\kern2\pthgap}\or
  \or\setbox\pta=\hbox{\box\pta\kern\pthgap\box\ptb}
  \or\setbox\pta=\hbox{\box\pta\kern\pthgap\box\ptb\kern\pthgap\box\ptc}\fi
  \ptxx=0.5\wd\pta
  \setbox\ptx=\vtop{
    \hbox{\ptedge{0}{0}{\wd\pta}{}}\nointerlineskip
    \box\pta}}

\newcount\pted          
\newcount\ptedm         
\newcount\pteda         
\newcount\ptedb         
\newcount\ptedc         
\newcount\ptedl         
\newcount\ptedh         
\newcount\ptedhs        
\newcount\ptedvs        
\newcount\ptedtemp      

\def\ptedge#1#2#3#4{\pted=#1%
  \pteda=#2\ifcase\pted\ptedb=#3\or\or\ptedb=#3\or\ptedb=#3\ptedc=#4\fi
  \ptedm=\pteda\advance\ptedm\ifcase\pted\ptedb\or\pteda\or\ptedb\or\ptedc\fi
  \divide\ptedm by 2
  \ptedh=\ptvgap\ptedtemp=\ptedm\advance\ptedtemp-\pteda\divide\ptedtemp by 6
  \ifnum\ptedh<\ptedtemp\ptedh=\ptedtemp\fi
  \unitlength=1sp%
  \begin{picture}(0,\ptedh)
    \ifnum\pted=3 \ptedput\ptedc\fi
    \ifnum\pted=1 \else\ptedput\ptedb\fi
    \ptedput\pteda
    \ifnum\pted=0 \ptedbot\fi 
  \end{picture}}

\def\ptedput#1{\ptedl=#1\advance\ptedl-\ptedm
  \ifnum\ptedl>0 \ptedslope\else
    \ptedl=-\ptedl\ptedslope\ptedhs=-\ptedhs\fi
  \ifnum\ptedhs=0 \ptedl=\ptedh\fi
  \put(\ptedm,\ptedh){\line(\ptedhs,-\ptedvs){\ptedl}}}

\def\ptedbot
 {\ptedtemp=\ptedl\multiply\ptedtemp by \ptedvs\divide\ptedtemp by \ptedhs
  \ifnum\ptedtemp>0\ptedtemp=-\ptedtemp\fi \advance\ptedtemp-1
  \advance\ptedtemp\ptedh\multiply\ptedl by 2
  \put(\pteda,\ptedtemp){\line(1,0){\ptedl}}}

\def\ptedslope
 {\ifnum\ptedl>\ptedh\ptedhs=\ptedl\ptedvs=\ptedh
  \else              \ptedvs=\ptedl\ptedhs=\ptedh \fi
  \divide\ptedhs by 60 \divide\ptedvs by \ptedhs
  \ifnum \ptedvs <  5 \ptedvs=0 \ptedhs=1 \else
  \ifnum \ptedvs < 11 \ptedvs=1 \ptedhs=6 \else
  \ifnum \ptedvs < 13 \ptedvs=1 \ptedhs=5 \else
  \ifnum \ptedvs < 17 \ptedvs=1 \ptedhs=4 \else
  \ifnum \ptedvs < 22 \ptedvs=1 \ptedhs=3 \else
  \ifnum \ptedvs < 27 \ptedvs=2 \ptedhs=5 \else
  \ifnum \ptedvs < 33 \ptedvs=1 \ptedhs=2 \else
  \ifnum \ptedvs < 38 \ptedvs=3 \ptedhs=5 \else
  \ifnum \ptedvs < 42 \ptedvs=2 \ptedhs=3 \else
  \ifnum \ptedvs < 46 \ptedvs=3 \ptedhs=4 \else
  \ifnum \ptedvs < 49 \ptedvs=4 \ptedhs=5 \else
  \ifnum \ptedvs < 55 \ptedvs=5 \ptedhs=6 \else
                      \ptedvs=1 \ptedhs=1 
    \fi \fi \fi \fi \fi \fi \fi \fi \fi \fi \fi \fi
  \ifnum \ptedl < \ptedh
    \ptedtemp=\ptedhs \ptedhs=\ptedvs \ptedvs=\ptedtemp \fi}

\newenvironment{parsetree}{\ptactivechardefs\ptbegtree}{\ptendtree}

\def\ptcatcodes
 {\catcode`(=\active \catcode`)=\active
  \catcode`.=\active \catcode``=\active
  \catcode`~=\active}

{\ptcatcodes
\gdef\ptactivechardefs
 {\ptcatcodes
  \def({\ptbeg\ignorespaces}
  \def){\ptend\ignorespaces}
  \def.##1.{\ptnode{##1}\ignorespaces} 
  \def`##1'{\ptleaf{##1}\ignorespaces}
  \def~{\pttritrue\ignorespaces}}}

\title{\vspace{-0.5in}Connectivity in Bag Generation}
\author{Arturo Trujillo and Simon Berry\thanks{\ Now at SHARP Laboratories of
Europe, Oxford Science Park, Oxford OX4 4GA. E-mail: simon@sharp.co.uk}\\
School of Computer and Mathematical Sciences \\ 
The Robert Gordon University, St Andrew Street \\
Aberdeen AB25 1HG \\
Scotland\\
\{iat,cs5sby\}@scms.rgu.ac.uk}

\begin{document}

\maketitle
\vspace{-0.5in}
\begin{abstract}

This paper presents a pruning technique which can be used to reduce
the number of paths searched in rule-based bag generators of the type
proposed by \cite{poznanskietal95} and \cite{popowich95}.  Pruning the
search space in these generators is important given the computational
cost of bag generation. The technique relies on a connectivity
constraint between the semantic indices associated with each lexical
sign in a bag. Testing the algorithm on a range of sentences shows
reductions in the generation time and the number of edges constructed.

\end{abstract}

\section{Introduction}

Bag generation is a form of natural language generation in which the
input is a bag (also known as a multiset: a set in which repeated
elements are significant) of lexical elements and the output is a
grammatical sentence or a statistically most probable permutation with
respect to some language model. 

Bag generation has been considered within the statistical and
rule-based paradigms of computational linguistics, and each has
handled this problem differently
\cite{chenetal94,whitelock94,popowich95,trujillo95c}. This
paper only considers rule based approaches to this problem.

Bag generation has received particular attention in lexicalist 
approaches to MT, as exemplified by Shake-and-Bake generation
\cite{beaven92,whitelock94}.
One can also envisage applications of bag generation to generation
from minimally recursive semantic representations
\cite{copestakeetal95b} and other semantic frameworks which separate scoping
from content information \cite{reyle95}.  In these frameworks, the
unordered nature of predicate or relation sets makes the application
of bag generation techniques attractive.

A notational convention used in the paper is that items such as `dog$_{1}$'
stand for simplified lexical signs of the form \cite{shieber86}:
\begin{quote}
{\scriptsize
$\avmplus{
\attvaltyp{cat}{N}\\
\attval{sem}{\avmplus{
		\attvaltyp{reln}{dog}\\
		\attvaltyp{arg1}{1}}}}$}
\end{quote}
In such signs, the semantic argument will be referred to as an `index' and
will be shown as a subscript to a lexeme;
in the above example, the index has been given the unique type {\bf 1}.

The term index is borrowed from HPSG \cite{pollardetal94} where
indices are used as arguments to relations; however these indices may
also be equated with discourse referents in DRT \cite{kampetal93}.  As
with most lexicalist generators, semantic variables must be
distinguished in order to disallow translationally incorrect
permutations of the target bag.
We distinguish variables by uniquely typing them.

Two assumptions are made regarding lexical-semantic indexing. 
\begin{assume}
All lexical signs must be indexed, including functional and
nonpredicative elements \cite{calderetal89}.
\end{assume}
\begin{assume}
All lexical signs must be connected to each other. Two lexical signs
are connected if they are directly connected; furthermore, the
connectivity relation is transitive. 
\label{con-con-ass}
\end{assume}
\begin{definition}
Two signs, A, B, are directly connected if there exist at
least two paths, PathA, PathB, such that A:PathA is token identical
with B:PathB.
\end{definition}

The indices involved in determining connectivity are
specified as parameters for a particular formalism. For example, in
HPSG, they would be indicated through paths such as {\sc
synsem:local:content:index}.

To ensure that only connected lexical signs are generated
and analysed, the following assumption must also be made:
\begin{assume}
A grammar will only generate or analyse connected lexical signs.
\label{con-gra-ass}
\end{assume}

\section{Bag Generation Algorithms}
\label{sb-alg-sec}

Two main types of rule-based bag generators have been proposed. The
first type consists of a parser suitably relaxed to take into account
the unordered character of the input
\cite{whitelock94,popowich95,trujillo95c}. For example, in
generators based on a chart parser, the fundamental rule 
is applied only when the edges to be combined share no lexical leaves,
in contrast to requiring that the two edges have source and target nodes
in common. The other type of generator applies a greedy algorithm to
an initial solution in order to find a grammatical sentence
\cite{poznanskietal95}. 

\subsection{Redundancy in Bag Generation}

One disadvantage with the above generators is that they construct a
number of structures which need not have been computed at all. In building
these structures, the generator is effectively searching branches of the
search space which never lead to a complete sentence. Consider the
the following input bag:
\begin{exquote}
{\em \{dog,barked,the,brown,big\}}
\end{exquote}
Previous researchers \cite{brew92,phillips93} have noted that from
such a bag, the following strings are generated but none can form part
of a complete sentence (note that indices are omitted when there is no
possibility of confusion; \# indicates that the substring will never
be part of a complete sentence):
\begin{example}
\# the dog\\
\# the dog barked\\
\# the brown dog
\label{two-red-exa}
\end{example}
For simple cases in chart based generators such unnecessary strings
do not create many problems, but for longer sentences, each additional
substring implies a further branch in the search tree
to be considered.

Since the computational complexity of the greedy bag generator
\cite{poznanskietal95} is polynomial (i.e. $\mathcal{O}$$(n^{4})$),
the effect of redundant substructures is not as detrimental as for
parser based generators. Nevertheless, a certain amount of unnecessary
work is performed. To show this, consider the test-rewrite sequence
for Example \ref{two-red-exa}:

\begin{sdescription}
\item{\bf Test:} dog barked the brown big 
\item{\bf Rewrite:} \_\_ barked the {\bf dog} brown big
\item{\bf Test:} barked (the dog) brown big 
\item{\bf Rewrite:} \_\_ (the dog) {\bf barked} brown big 
\item{\bf Test:} ((the dog) barked) brown big 
\item{\bf Rewrite:} the {\bf brown} dog barked \_\_ big 
\item{\bf Test:} ((the (brown dog)) barked) big 
\item{\bf Rewrite:} the {\bf big} (brown dog) barked \_\_ 
\item{\bf Test:} ((the (big (brown dog))) barked) {\em (terminate)}
\end{sdescription}

In this sequence double underscore
(\_\_) indicates the starting position of a moved constituent; the moved
constituent itself is given in bold face; the bracketing indicates
analysed constituents (for expository purposes the algorithm has been
oversimplified, but the general idea remains the same).

Now consider the step where `brown' is inserted between `the' and
`dog'. This action causes the complete structure for `the dog barked'
to be discarded and replaced with that for `the brown dog barked',
which in turn is discarded and replaced by `the big brown dog barked'.

\subsection{Previous Work}

A number of pruning techniques have been suggested
to reduce the amount of redundancy in bag generators.
Brew \shortcite{brew92}
proposed a constraint propagation technique which eliminates
branches during bag generation by considering the necessary
functor-argument relationships that exist between the component
basic signs of categorial signs. These relationships form a graph 
indicating the necessary conditions for a lexical item to form part of
a complete sentence.
Such graphs can be used to eliminate the substrings in Example
\ref{two-red-exa}.  Unfortunately the technique exploits
specific aspects of categorial grammars and it is not clear how they
may be used with other formalisms.

Trujillo \shortcite{trujillo95c} adapts some of Brew's ideas to phrase
structure grammars by compiling Follow functions and constructing
adjacency graphs. While this approach reduces the size of the search
space, it does not prune it sufficiently for certain classes of
modifiers.

Phillips \shortcite{phillips93} proposes handling inefficiency at the
expense of completeness. His idea is to maintain a queue of modifiable
constituents (e.g. N1s) in order to delay their combination with other
constituents until modifiers (e.g. PPs) have been analysed.  While
practical, this approach can lead to alternative valid sentences not
being generated.

\section{Connectivity Restrictions}

In searching for a mechanism that eliminates unnecessary wfss,
it will be possible to use indices in lexical signs.
As mentioned earlier, these indices play a major role in preventing the
generation of incorrect translations.

\begin{figure}[htbp]
{\scriptsize
1) $\avmplus{
\attvaltyp{cat}{S}\\
\attval{sem}{\ind{0}}}
\Longrightarrow
\avmplus{
\attvaltyp{cat}{NP}\\
\attval{sem:arg1}{\ind{1}}}
\hspace{0.5em}
\avmplus{
\attvaltyp{cat}{VP}\\
\attval{sem}{\ind{0}}\avmplus{
			\attval{arg2}{\ind{1}}}}$ \smallnl

2) $\avmplus{
\attvaltyp{cat}{NP}\\
\attval{sem}{\ind{0}}}
\Longrightarrow
\avmplus{
\attvaltyp{cat}{Det}\\
\attval{sem:arg1}{\ind{1}}}
\hspace{0.5em}
\avmplus{
\attvaltyp{cat}{N1}\\
\attval{sem}{\ind{0}}\avmplus{
			\attval{arg1}{\ind{1}}}}$ \smallnl

3) $\avmplus{
\attvaltyp{cat}{N1}\\
\attval{sem}{\ind{0}}}
\Longrightarrow
\avmplus{
\attvaltyp{cat}{A}\\
\attval{sem:arg1}{\ind{1}}}
\hspace{0.5em}
\avmplus{
\attvaltyp{cat}{N1}\\
\attval{sem}{\ind{0}}\avmplus{
			\attval{arg1}{\ind{1}}}}$ \smallnl

4) $\avmplus{
\attvaltyp{cat}{N1}\\
\attval{sem}{\ind{0}}}
\Longrightarrow
\avmplus{
\attvaltyp{cat}{N1}\\
\attval{sem:arg1}{\ind{1}}}
\hspace{0.5em}
\avmplus{
\attvaltyp{cat}{PP}\\
\attval{sem}{\ind{0}}\avmplus{
			\attval{arg1}{\ind{1}}}}$ \smallnl

5) $\avmplus{
\attvaltyp{cat}{N1}\\
\attval{sem}{\ind{0}}}
\Longrightarrow
\avmplus{
\attvaltyp{cat}{N}\\
\attval{sem}{\ind{0}}}$ \smallnl

6) $\avmplus{
\attvaltyp{cat}{PP}\\
\attval{sem}{\ind{0}}}
\Longrightarrow
\avmplus{
\attvaltyp{cat}{P}\\
\attval{sem}{\ind{0}}\avmplus{
			\attval{arg3}{\ind{2}}}}
\hspace{0.5em}
\avmplus{
\attvaltyp{cat}{NP}\\
\attval{sem:arg1}{\ind{2}}}$ \smallnl

7) $\avmplus{
\attvaltyp{cat}{VP}\\
\attval{sem}{\ind{0}}}
\Longrightarrow
\avmplus{
\attvaltyp{cat}{Vtra}\\
\attval{sem}{\ind{0}}\avmplus{
			\attval{arg3}{\ind{2}}}} 
\hspace{0.5em}
\avmplus{
\attvaltyp{cat}{NP}\\
\attval{sem:arg1}{\ind{2}}}$}
\caption{Simple unification grammar.}
\label{sim-gra-fig}
\end{figure}

It will be shown that it is possible to exploit the connectivity
Assumption \ref{con-con-ass} above in order to achieve a reduction in
the number of redundant wfss constructed by both types of generator
described in section \ref{sb-alg-sec}.

\subsection{Using Connectivity for Pruning}

Take the following bag:
\begin{example}
\{dog$_{1}$,the$_{1}$,brown$_{1}$,big$_{1}$\}
\label{aan-bag-exa}
\end{example}
(corresponding to `the big brown dog').  Assume that the next wfss to
be constructed by the generator is the NP `the dog'. Given the grammar
in Figure \ref{sim-gra-fig}, it is possible to deduce that `brown' can
never be part of a complete NP constructed from such a substring.
This can be determined as follows. If this adjective were part of such
a sentence, `brown' would have to appear as a leaf in some constituent
that combines with `the dog' or with a constituent containing `the
dog'. From the grammar, the only constituents that can combine with
`dog' are VP, Vtra and P. However, none of these constituents can have
`brown$_{1}$' as a leaf: in the case of P and Vtra this is trivial,
since they are both categories of a different lexical type. In the
case of the VP, `brown$_{1}$' cannot appear as a leaf either because
expansions of the VP are restricted to NP complements with 2 as their
semantic index, which in turn would also require adjectives within
them to have this index. Furthermore, `brown$_{1}$' cannot occur as a
leaf in a deeper constituent in the VP because such an occurrence
would be associated with a different index.
In such cases `brown' would modify a different noun with a different index:
\begin{example}
\{the$_{1}$,dog$_{1}$,with$_{1,2}$,the$_{2}$,brown$_{2}$,collar$_{2}$\}
\end{example}

A naive implementation of this deduction would attempt to expand the
VP depth-first, left to right, in order to accommodate `brown' in a
complete derivation.  Since this would not be possible, the NP `the
dog' would be discarded.  This approach is grossly inefficient
however. What is required is a more tractable algorithm which, given a
wfss and its associated sign, will be able to determine whether all
remaining lexical elements can ever form part of a complete sentence
which includes that wfss.

Note that deciding whether a lexical sign can appear outside a phrase is
determined purely by the grammar, and not by whether the lexical
elements share the same index or not. Thus, a more complex grammar
would allow `the man' from the bag
\begin{example}
\{the$_{1}$,man$_{1}$,shaves$_{e,1,1}$,himself$_{1}$\}
\end{example}
even though `himself' has the same index as `the man'. 

\subsection{Outer Domains}

The approach introduced here compiles the relevant information offline
from the grammar and uses it to check for connectivity during bag
generation. The compilation process results in a set of
(Sign,Lex,Bindings) triples called {\em outer domains}. This set is
based on a unification-based phrase structure grammar defined as follows:
\begin{definition}
A grammar is a tuple (N,T,P,S), where P is a set of productions
$\alpha \Rightarrow \beta$, $\alpha$ is a sign, $\beta$ is a
list of signs, N is the set of all $\alpha$, T is the set of all signs
appearing as elements of $\beta$ which unify with lexical
entries, and S is the start sign.
\end{definition}
Outer domains are defined as follow:
\begin{definition}
\{ (Sign,Lex,Binds) $\mid$ Sign $\in N \cup T$, Lex $\in T$ and there
exists a derivation $\alpha \stackrel{*}{\Rightarrow} 
\beta_{1}$Sign$'$$\beta_{2}$Lex$'$$\beta_{3}$
or $\alpha \stackrel{*}{\Rightarrow}
\beta_{1}$Lex$'$$\beta_{2}$Sign$'$$\beta_{3}$,
and Sign$'$ a unifier
for Sign, Lex$'$ a unifier for Lex, and Binds the set of all path
pairs $<$SignPath,LexPath$>$ such that Sign$'$:SignPath is token
identical with Lex$'$:LexPath\}
\end{definition}
Intuitively, the outer domains indicate that preterminal category Lex
can appear in a complete sentence with subconstituent Sign, such that
Lex is not a leaf of Sign. Using ideas from data flow analysis
\cite{kennedy81}, predictive parser constructions \cite{ahoetal86} and
feature grammar compilation \cite{trujillo94} it is
possible to construct such a set of triples.  Outer domains thus
represent elements which may lie outside a subtree of category Sign in
a complete sentential derivation. The following definition specifies
how outer domains are used:
\begin{definition}
A lexical sign Lex$'$ is in the outer domain of Sign$'$ iff there is a triple
(Sign,Lex,Binds) in outer domains such that Sign and Lex unify with 
Sign$'$ and Lex$'$ respectively, and there is at least one pair 
$<$PathS,PathL$>$ $\in$ Binds such that Sign$'$:PathS unifies
with Lex$'$:PathL.
\end{definition}

In compiling outer domains, {\em inner domains} are used to facilitate
computation.  Inner domains are defined as follows:
\begin{definition}
\{ (Sign,Lex,Binds) $\mid$ Sign $\in N \cup T$, Lex $\in T$ and there
exists a derivation $\alpha \stackrel{*}{\Rightarrow}
\beta_{1}$Lex$'$$\beta_{2}$,
with Sign$'$ a unifier for Sign, Lex$'$ a unifier
for Lex, and Binds the set of all path pairs $<$SignPath,LexPath$>$
such that Sign$'$:SignPath is token identical with Lex$'$:LexPath\}
\end{definition}
The inner domains thus express all the possible terminal categories
which may be derived from each nonterminal in the grammar. 

To be able to exploit connectivity during generation, inner and outer
domains contain only triples in which Binds has at least one element.
In this way, only those lexical categories which are directly
connected to the sign are taken into account; the implication of this
will become clearer later.

As an example, the outer domain of NP as derived from the above grammar is:
\begin{exquote}
(NP[sem:arg1:X],Vtra[sem:arg2:Y],\\
\tab \{$<$sem:arg1,sem:arg2$>$\})\\
(NP[sem:arg1:X],Vtra[sem:arg3:Y],\\
\tab \{$<$sem:arg1,sem:arg3$>$\})\\
(NP[sem:arg1:X],P[sem:arg3:Y],\\
\tab \{$<$sem:arg1,sem:arg3$>$\})
\end{exquote}
This set indicates that for any NP, the only terminal categories not
contained in the subtree with root NP, and with which the NP shares a
semantic index, are Vtra and P. For instance, the first triple arises
from the following tree:

{\scriptsize
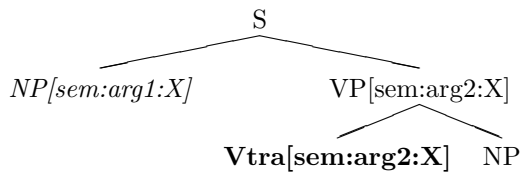
\begin{parsetree}
(.S. `NP[sem:arg1:X]'
     (.VP[sem:arg2:X]. `{\bf Vtra[sem:arg2:X]}'
		       `{\rm NP}')) 
\end{parsetree}
}

\subsection{Pruning through Outer Domains and Connectivity}

The pruning technique developed here operates on grammars whose
analyses result in connected leaves. 

Consider some wfss W constructed from a bag B and with category C;
this category, in the form of a sign, will include syntactic and
lexical-semantic information.  Such a wfss will have been constructed
during the bag generation process. Now, either W includes all the
input elements as leaves, in which case W constitutes a complete
sentence, or there are elements in the input bag which are not part of
W. In the latter case, for bags obeying Assumption \ref{con-con-ass},
the following condition holds for any W that can form part of a
complete sentence:
\begin{condition}
Let L be the set of leaves appearing in W, let G be the graph (V,E),
where V = \{C\} $\cup$ B $-$ L, and E = \{ \{x,y\} $|$ x,y $\in$ V and
y is in the outer domain of x\}. Then G is connected.
\label{con-gra-con}
\end{condition}

To show that this condition indeed holds, consider a grammatical
ordering of some input bag B, represented as the string W:
\begin{exquote}
$\alpha$..$\gamma\delta$..$\omega$
\end{exquote}
By Assumption \ref{con-con-ass}, the lexical elements in the bag, and
therefore in any grammatical ordering of it, are connected. Now consider
reducing this string using the production rule:
\begin{exquote}
D $\Rightarrow \gamma\delta$
\end{exquote}
to give the string W$'$:
\begin{exquote}
$\alpha$..D..$\omega$
\end{exquote}
In this case, the signs in W$'$ will also be connected. This can be
shown by contradiction:
\begin{proof}
Assume that there is some sign $\zeta$ in W$'$ to which D is not
connected. Then grammar G would allow disconnected strings to be
generated, contrary to Assumption \ref{con-gra-ass}. This is because D
would not be able to rewrite
$\gamma_{1}\delta_{1}$ in
such a way that both
daughters were connected to $\zeta$, leading to a disconnected
string.
\end{proof}

The situation in string W$'$ is analogous to that in Condition
\ref{con-gra-con}. By identifying signs which are directly connected
in E, it is possible to determine whether E is connected and
consequently whether C can form part of a complete derivation. Instead
of simply comparing the value of index paths, it is more restrictive
to use outer domains since they give us precisely those elements
which are directly connected to a sign and are in its outer domain.

\subsection{Example}

\begin{figure}[htbp] \centering
\setlength{\unitlength}{0.00083300in}%
\begingroup\makeatletter\ifx\SetFigFont\undefined
\def\x#1#2#3#4#5#6#7\relax{\def\x{#1#2#3#4#5#6}}%
\expandafter\x\fmtname xxxxxx\relax \def\y{splain}%
\ifx\x\y   
\gdef\SetFigFont#1#2#3{%
  \ifnum #1<17\tiny\else \ifnum #1<20\small\else
  \ifnum #1<24\normalsize\else \ifnum #1<29\large\else
  \ifnum #1<34\Large\else \ifnum #1<41\LARGE\else
     \huge\fi\fi\fi\fi\fi\fi
  \csname #3\endcsname}%
\else
\gdef\SetFigFont#1#2#3{\begingroup
  \count@#1\relax \ifnum 25<\count@\count@25\fi
  \def\x{\endgroup\@setsize\SetFigFont{#2pt}}%
  \expandafter\x
    \csname \romannumeral\the\count@ pt\expandafter\endcsname
    \csname @\romannumeral\the\count@ pt\endcsname
  \csname #3\endcsname}%
\fi
\fi\endgroup
\begin{picture}(2208,1767)(4426,-2422)
\thicklines
\put(6601,-961){\vector( 3, 2){  0}}
\put(6601,-961){\vector(-3,-2){1800}}
\put(4801,-961){\vector(-3, 2){  0}}
\put(4801,-961){\vector( 3,-2){1800}}
\put(6601,-961){\vector( 0, 1){  0}}
\put(6601,-961){\vector( 0,-1){1200}}
\put(4801,-961){\vector(-1, 0){  0}}
\put(4801,-961){\vector( 1, 0){1800}}
\put(4801,-961){\vector( 0, 1){  0}}
\put(4801,-961){\vector( 0,-1){1200}}
\put(6601,-2161){\vector( 1, 0){  0}}
\put(6601,-2161){\vector(-1, 0){1800}}
\put(6601,-886){\makebox(0,0)[lb]{\smash{\SetFigFont{12}{14.4}{rm}the$_{1}$}}}
\put(4501,-2311){\makebox(0,0)[lb]{\smash{\SetFigFont{12}{14.4}{rm}big$_{1}$}}}
\put(6526,-2386){\makebox(0,0)[lb]{\smash{\SetFigFont{12}{14.4}{rm}brown$_{1}$}}}
\put(4426,-811){\makebox(0,0)[lb]{\smash{\SetFigFont{12}{14.4}{rm}dog$_{1}$}}}
\end{picture}

\caption{Initial connected graph.}
\label{ini-con-fig}
\end{figure}
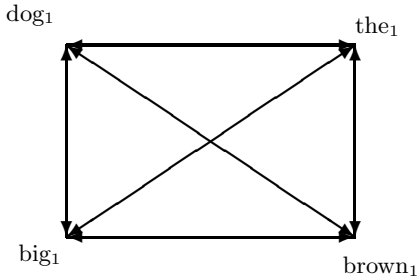

Consider Example \ref{aan-bag-exa}. To eliminate the wfss `the dog'
from further consideration, a connected graph of lexical signs is
constructed before generation is started (Figure \ref{ini-con-fig}).
This graph is built by using the outer domain of each lexical element
to decide which of the remaining elements could possibly share an index
with it in a complete sentence.

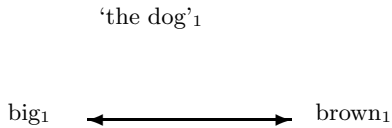
\begin{figure}[htbp] \centering
\setlength{\unitlength}{0.00083300in}%
\begingroup\makeatletter\ifx\SetFigFont\undefined
\def\x#1#2#3#4#5#6#7\relax{\def\x{#1#2#3#4#5#6}}%
\expandafter\x\fmtname xxxxxx\relax \def\y{splain}%
\ifx\x\y   
\gdef\SetFigFont#1#2#3{%
  \ifnum #1<17\tiny\else \ifnum #1<20\small\else
  \ifnum #1<24\normalsize\else \ifnum #1<29\large\else
  \ifnum #1<34\Large\else \ifnum #1<41\LARGE\else
     \huge\fi\fi\fi\fi\fi\fi
  \csname #3\endcsname}%
\else
\gdef\SetFigFont#1#2#3{\begingroup
  \count@#1\relax \ifnum 25<\count@\count@25\fi
  \def\x{\endgroup\@setsize\SetFigFont{#2pt}}%
  \expandafter\x
    \csname \romannumeral\the\count@ pt\expandafter\endcsname
    \csname @\romannumeral\the\count@ pt\endcsname
  \csname #3\endcsname}%
\fi
\fi\endgroup
\begin{picture}(2025,795)(4801,-1600)
\thicklines
\put(5401,-1561){\vector(-1, 0){  0}}
\put(5401,-1561){\vector( 1, 0){1275}}
\put(5476,-961){\makebox(0,0)[lb]{\smash{\SetFigFont{12}{14.4}{rm}`the dog'$_{1}$}}}
\put(6826,-1561){\makebox(0,0)[lb]{\smash{\SetFigFont{12}{14.4}{rm}brown$_{1}$}}}
\put(4906,-1561){\makebox(0,0)[lb]{\smash{\SetFigFont{12}{14.4}{rm}big$_{1}$}}}
\end{picture}

\caption{Updated disconnected graph after the wfss `the dog' is constructed.}
\label{upd-gra-fig}
\end{figure}

When a new wfss is constructed during generation, say by application
of the modified fundamental rule or during the rewrite phase in a
greedy algorithm, this initial graph is updated and tested for
connectivity. If the updated graph is not connected then the proposed
wfss cannot form part of a complete sentence.  Updating the graph
involves three steps. Firstly every node in the graph which is a leaf
of the new wfss is deleted, together with its associated
arcs. Secondly, a new node corresponding to the new wfss is added to
the graph. Finally, a new arc is added to the graph between the new
node and every other node lying in its outer domain.  The updated
(disconnected) graph that ensues after constructing `the dog' is
shown in Figure \ref{upd-gra-fig}; this NP is therefore rejected.

\section{Compiling Connectivity Domains}

For reasons of space, the computation of outer domains cannot be
described fully here. The broad outline, however, is as
follows. First, the inner domains of the grammar are calculated. This
involves the calculation of the fixed point of set equations,
analogous to those used in the construction of First sets for
predictive parsers \cite{ahoetal86,trujillo94}. Given the
inner domains of each category in the grammar, the construction of the
outer domains involves the computation of the fixed point of set
equations relating the outer domain of a category to the inner domain
of its sisters and to the outer domain of its mother, in a manner
analogous to the computation of Follow sets.

During computation, the set of Binds is monotonically increased
as different ways of directly connecting sign and lexeme are found.

\section{Results}

The above pruning technique has been tested on bags of different sizes
including different combinations of modifiers. Sentences were
generated using two versions of a modified chart parser. In one, every
inactive edge constructed was added to the chart. In the other, every
inactive edge was tested to see if it led to a disconnected graph; if
it did, then the edge was discarded. The results of the experiment are
shown in Table \ref{res-com-tab}.
The implementation was in Prolog
on a Sun SparcStation 10; the generation timings do not
include garbage collection time. The grammar
used for the experiment consisted of simplified,
feature-based versions of the ID rules in GPSG; there were 18 rules
and 50 lexical entries. Compilation of the outer domains for these
rules took approximately 37 minutes, and the resulting set occupies 40K of
memory. In the general case, however, the size of the outer domains is
$\mathcal{O}$$(n^{2})$, where $n$ is the number of distinct signs;
this number can be controlled by employing equivalence classes of
different levels of specificity for
pre-terminal and non-terminal signs.
\begin{table}[htbp] \centering
\small
\begin{tabular}{lrrrr} \cline{2-5}
\multicolumn{1}{c}{} & \multicolumn{2}{c}{\em Chart Gen.} & \multicolumn{2}{c}{\em + Pruning} \\ \hline
{\em Bag size} & {\em Time} & {\em Edges} & {\em Time} & {\em Edges} \\ \hline
2  & 0.1       & 15   &  0.1 & 15 \\
4 & 0.3 & 37 & 0.4 & 36 \\
7 & 1.5 & 103 & 2.0 & 99 \\
7 & 0.9 & 72 & 1.0 & 67 \\
11 & 5.1 & 213 & 3.9 & 138 \\
12 & 2.6 & 133 & 3.4 & 123\\
15 & 9.0 & 294 & 7.2 & 186 \\
15 & 17.6 & 448 & 11.1 & 253 \\
17 & 2.3 & 126 & 2.6 & 105 \\ \hline
\end{tabular}
\caption{Effect of pruning (times in secs).}
\label{res-com-tab}
\end{table}

Only one reading was generated for each bag, corresponding to one
attachment site for PPs. The table shows that the technique can yield
reductions in the number of edges (both active and inactive)
and time taken, especially for longer sentences, while retaining
the overheads at an acceptable level.

\section{Conclusion}

A technique for pruning the search space of a bag generator has been
implemented and its usefulness shown in the generation of
different types of constructions. The technique relies on a
connectivity constraint imposed on the semantic relationships
expressed in the input bag. In order to apply the algorithm, outer
domains needed to be compiled from the grammar; these are used to
discard wfss by ensuring lexical signs outside a wfss can indeed appear
outside that string.

Exploratory work employing adjacency constraints during generation has
yielded further improvements in execution time when applied in
conjunction with the pruner. If extended appropriately, these
constraints could prune the search space even further.
This work will be reported at a later date.

\section*{Acknowledgments}

Two anonymous reviewers provided very useful comments; 
we regret not being able to do justice to all their suggestions.


\end{document}